\documentclass[letterpaper, 11 pt]{article} 
\usepackage{amsmath,amsfonts,amssymb,color}
\usepackage{mathtools}
\usepackage{dsfont}
\usepackage[dvipsnames]{xcolor}
\definecolor{mygreen}{rgb}{0.0, 0.5, 0.0}
\definecolor{winered}{rgb}{0.8,0,0}
\definecolor{myblue}{rgb}{0,0,0.8}
\usepackage{tikz}
\usepackage{amsthm}
\usepackage{empheq}

\newtheorem{definition}{Definition}
\newtheorem{theorem}{Theorem}
\newtheorem{lemma}{Lemma}

\newtheorem{corollary}{Corollary}
\newtheorem{remark}{Remark}
\newtheorem{assumption}{Assumption}

\newcommand{\tb}{\color{blue}}
\newcommand{\tr}{\color{red}}
\newcommand\x{{\boldsymbol{\theta}}}
\newcommand\xk{{\boldsymbol{\theta}_k}}
\newcommand\xktau{{\boldsymbol{\theta}_{k-\tau}}}
\newcommand\xs{{\boldsymbol{\theta}^*}}
\newcommand\xkp{{\boldsymbol{\theta}_{k+1}}}

\newcommand\dx{{\|\xs - \xk\|^2}}
\newcommand\dxpd{{\delta_{k+1}^2}}
\newcommand\dxd{{\delta_k^2}}
\newcommand\dxdtau{{\delta_{k-\tau}^2}}
\newcommand\sumN{\frac{1}{N}\sum_{i = 1}^N}
\newcommand{\myFrac}[2]{\frac{#1}{#2}}
\newcommand\dxddifftau{{\delta_{k,\tau}^2}}

\newcommand\dxzd{{\delta_0^2}}

\newcommand\vk{\mathbf{v}_k}

\newcommand\gkn{\mathbf{g}_{i,k}}
\newcommand\gkm{\mathbf{g}_{j, k}}

\newcommand\bkn{b_{i,k}}
\newcommand\bkm{b_{j,k}}
\newcommand\hkn{\mathbf{h}_{i,k}}

\newcommand\okn{o_{i, k}}

\newcommand\okntau{o_{i, k-\tau}}
\newcommand\okmtau{o_{j, k-\tau}}
\newcommand\bfg{\mathbf{g}}

\newcommand\barg{\bar{\mathbf{g}}}

\newcommand\muiktau{\eta_{k, \tau}^{(i)}}
\newcommand\mujktau{\eta_{k, \tau}^{(j)}}

\newcommand{\E}[1]{\mathbb{E}\left[#1\right]}
\newcommand{\eqal}[2]{\begin{equation}\begin{aligned}\label{#1}
			#2
		\end{aligned}\end{equation}}
\newcommand{\eqalNo}[2]{\begin{equation*}\begin{aligned}\label{#1}
			#2
		\end{aligned}\end{equation*}}
\usepackage{cancel}

\usepackage{algorithm}
\usepackage{algpseudocode}
\usepackage{epsfig}
\usepackage{array}
\usepackage{multirow}
\usepackage{epstopdf}
\usepackage{tikz}
\usepackage{relsize}
\usetikzlibrary{shapes,arrows}
\usepackage[hidelinks]{hyperref}
\hypersetup{
    colorlinks=true,
    linkcolor={winered},
    citecolor={mygreen}
}
\usepackage{cite}
\usepackage[margin=1in]{geometry}

\begin{document}
\title{\LARGE  
	Federated TD Learning over Finite-Rate   
 Erasure Channels: \\ Linear Speedup under Markovian Sampling}

\author{Nicol\`o Dal Fabbro, Aritra Mitra and George J. Pappas
	\thanks{N. Dal Fabbro is with the Department of Information Engineering, University of Padova. Email: dalfabbron@dei.unipd.it. A. Mitra is with the Electrical and Computer Engineering Department, North Carolina State University. Email: amitra2@ncsu.edu. G. J. Pappas is with the Electrical and Systems Engineering Department, University of Pennsylvania. Email: pappasg@seas.upenn.edu. This work was supported by NSF Award 1837253, and the Italian Ministry of Education, University and Research through the PRIN project no. 2017NS9FEY.}
}
\maketitle
\begin{abstract}
    Federated learning (FL) has recently gained much attention due to its effectiveness in speeding up supervised learning tasks under communication and privacy constraints. However, whether similar speedups can be established for reinforcement learning remains much less understood theoretically. Towards this direction, we study a federated policy evaluation problem where agents communicate via a central aggregator to expedite the evaluation of a common policy. To capture typical communication constraints in FL, we consider finite capacity up-link channels that can drop packets based on a Bernoulli erasure model. Given this setting, we propose and analyze \texttt{QFedTD} - a quantized federated temporal difference learning algorithm with linear function approximation. Our main technical contribution is to provide a finite-sample analysis of \texttt{QFedTD} that (i) highlights the effect of quantization and erasures on the convergence rate; and (ii) establishes a linear speedup w.r.t. the number of agents under Markovian sampling. Notably, while different quantization mechanisms and packet drop models have been extensively studied in the federated learning, distributed optimization, and networked control systems literature, our work is the first to provide a non-asymptotic analysis of their effects in multi-agent and federated reinforcement learning. 
\end{abstract}
\section{Introduction}
Is it possible to obtain statistical models of high accuracy for supervised learning problems (e.g., regression, classification, etc.) by aggregating information from multiple devices while keeping the raw data on these devices private? This is the central question of interest in the popular machine learning paradigm of federated learning (FL) \cite{konevcny, mcmahan, bonawitz}. When the data-generating distributions of the participating devices are identical (or sufficiently similar), several works have shown that one can reap the benefits of collaboration by exchanging locally trained models via a central aggregator (server) \cite{stich, khaled1, khaled2, haddadpour, woodworth1, scaffold, acar2021, gorbunov, mitraNIPS, proxskip, collinsfedavg}. In practice, these models are typically high-dimensional and need to be exchanged over unreliable communication links of limited bandwidth. As such, a large body of work in FL has investigated the effects of uploading quantized models (or model-differentials, i.e., gradients) over channels prone to packet drops/erasures \cite{FedPAQ, had21}. Drawing inspiration from this literature, in this paper, we ask: \textit{Can we establish collaborative performance gains for federated reinforcement learning (FRL) problems subject to similar communication challenges?} As it turns out, little to nothing is known about this question from a theoretical standpoint. 

Towards this direction, we study one of the most basic problems in RL, namely \textit{policy evaluation}, in a federated setting. Specifically, in our problem, $N$ agents, each of whom interacts with the same Markov Decision Process (MDP), communicate via a server to evaluate a fixed policy. While each agent can evaluate the policy on its own using Monte-Carlo sampling or temporal difference (TD) learning algorithms \cite{sutton1988learning, tsitsiklisroy}, the reason for communicating is the same as in the standard FL setting: \textit{to achieve an $N$-fold speedup in the sample-complexity of policy evaluation relative to when an agent acts alone}. In the recent survey paper on FRL \cite{qiFRL}, the authors mention that the goal of the FRL framework is to achieve such speedups while respecting privacy constraints, i.e., without revealing the raw data (states, actions, and rewards) of the agents. Relative to the FL setting, proving finite-time rates for FRL is significantly more challenging since we need to deal with temporally correlated Markovian samples. Indeed, even for the single-agent setting, finite-time rates under Markovian sampling have only recently been established \cite{bhandari2018, srikant2019finite, chenQ, 
 patil2023}. Works prior to these developments either provided a finite-time analysis under a restrictive i.i.d. sampling assumption~\cite{lakshmi, dalal}, or only came with asymptotic guarantees~\cite{tsitsiklisroy, borkarode}. For the multi-agent setting, almost all the prior works on TD learning make a restrictive i.i.d. sampling assumption \cite{doan, liuMARL}. The only two exceptions to this are the very recent papers \cite{khodadadian2022federated, han} that establish linear speedups under Markovian sampling; however, none of the above works consider any communication constraints. As such, establishing linear speedups in FRL under Markovian sampling and communication constraints remains largely unexplored. In this regard, our main contributions are as follows. 

\textbf{Contributions.} Our first contribution is to formulate a federated policy evaluation problem under two practical constraints on the communication channels: finite capacity and packet drops (lossy links). To capture these constraints, we propose and analyze \texttt{QFedTD} - a federated TD algorithm with linear function approximation where agents upload quantized TD update directions to the server over Bernoulli erasure channels \cite{hadjicostis, schenato}. While various quantization and erasure models have been extensively analyzed in the FL \cite{FedPAQ, had21}, distributed optimization \cite{rabbat, doanquant, reisizadehTSP, michelusi}, and networked control literature \cite{hadjicostis, schenato} for almost two decades, our work is the first to formally study their non-asymptotic effects in the context of multi-agent/federated RL. 

Our second and most significant contribution is to provide a rigorous non-asymptotic analysis of \texttt{QFedTD} that clearly highlights the effects of quantization and erasures, and establishes an $N$-fold linear speedup in sample-complexity relative to the single-agent setting. Since RL algorithms often require several samples to achieve acceptable accuracy, our speedup result under realistic communication models is of significant practical importance. We now comment on some of the highlights of our analysis relative to \cite{khodadadian2022federated} and \cite{han}. Our work crucially departs from both these papers in that, in addition to correlated Markovian samples, we need to contend with two other sources of randomness: one due to randomized quantization and the other due to the Bernoulli packet-dropping processes. Even in the absence of communication challenges, our analysis has the following key benefits. Unlike \cite{han}, our work does not require any projection step to ensure the boundedness of iterates. Moreover, compared to \cite{han}, and the analysis in \cite{khodadadian2022federated} that relies on Generalized Moreau Envelopes, our proof is significantly shorter and simpler. As a byproduct of this simpler analysis, we derive bounds that have a tighter linear dependence on the mixing time (consistent with the centralized setting) as opposed to the quadratic dependence in \cite{khodadadian2022federated, han}.\footnote{To be fair, we should point out that \cite{khodadadian2022federated} and \cite{han} look at somewhat more general updating schemes than us by allowing for the agents to perform multiple local updates in every communication round. Instead, we only consider one local step in our analysis. While performing more than one local step leads to a ``client-drift" effect \cite{scaffold, charles, mitraNIPS}, it is not clear to us whether/why such a drift effect should lead to sub-optimal dependencies on the mixing time.} In fact, the dependence of $O(\tau)$ in our variance bounds (where $\tau$ is the mixing time) is information-theoretically  optimal~\cite{nagaraj}. The other natural advantage of our simple proof template is that one can potentially build on it while trying to establish linear speedups for more involved RL settings.

\newpage
\section{System Model and Problem Formulation}
\label{sec:system_model}
We consider a setting involving $N$ agents, where all agents interact with the \textit{same} Markov Decision Process (MDP). Let us denote the shared MDP by $\mathcal{M} = (\mathcal{S}, \mathcal{A}, \mathcal{P}, \mathcal{R}, \gamma)$, where $\mathcal{S}$ is a finite state space
of size $n$, $\mathcal{A}$ is a finite action space, $\mathcal{P}$ is a set of action-dependent Markov transition kernels, $\mathcal{R}$ is a reward function, and $\gamma \in (0,1)$ is the discount factor. We are interested in a \textit{policy evaluation} (PE) problem where the agents exchange information via a central aggregator (server) to evaluate the value function associated with a 
 policy $\mu$. Here, the policy is a map from the states to the actions, i.e., $\mu:\mathcal{S}\rightarrow\mathcal{A}$. In what follows, we first briefly review some key concepts relevant to PE with function approximation. Then, we formally describe our communication model, objectives, and technical challenges. 

\textbf{Policy Evaluation with Linear Function Approximation.} The policy $\mu$ to be evaluated induces a Markov Reward Process (MRP) with transition matrix $\mathbb{P}_{\mu}$ and reward function $R_{\mu}:\mathcal{S} \rightarrow \mathbb{R}$. The purpose of PE is to evaluate the value function $\boldsymbol{V}_{\mu}(s)$ for each $s\in\mathcal{S}$, where $\boldsymbol{V}_{\mu}(s)$ is the discounted expected cumulative reward obtained by playing policy $\mu$ starting from initial state $s$. Formally, we have
\begin{equation}
	\boldsymbol{V}_{\mu}(s) = \mathbb{E}\left[\sum_{k = 0}^\infty \gamma^kR_{\mu}(s_k)|s_0 = s\right],
 \label{eqn:Value_func}
\end{equation}
where $s_k$ represents the state of the Markov chain at the discrete time-step $k$ under the action of the policy $\mu$. Our particular interest is in the RL setting where the Markov transition kernels and reward functions are \textit{unknown}. 


In several large-scale practical settings, the size $n$ of the state space $\mathcal{S}$ is large, thereby creating a major computational challenge. To work around this issue, we will resort to the popular idea of linear function approximation where $\boldsymbol{V}_{\mu}$ is approximated by vectors in a linear subspace of $\mathbb{R}^n$ spanned by a set of $m$ basis vectors $\{\boldsymbol{\phi}_\ell\}_{\ell \in [m]}$\footnote{Given a positive integer $m$, we use the notation $[m] = 1, ..., m$.}; importantly,  $m \ll n$. To be more precise, let us define the feature matrix $\boldsymbol{\Phi} \triangleq [\boldsymbol{\phi}_1, ..., \boldsymbol{\phi}_m] \in \mathbb{R}^{n \times m}$. Given a weight (model) vector $\boldsymbol{\theta} \in \mathbb{R}^m$, the parametric approximation $\hat{\boldsymbol{V}}_{\boldsymbol{\theta}}$ of $\boldsymbol{V}_\mu$ is then given by $\boldsymbol{V}(\boldsymbol{\theta}):=\hat{\boldsymbol{V}}_{\boldsymbol{\theta}} = \boldsymbol{\Phi}\boldsymbol{\theta}.$ If we denote the $s$-th row of $\boldsymbol{\Phi}$ as ${\boldsymbol{{\phi}}}_s'$, then the approximation of $\boldsymbol{V}_{\mu}(s)$, in particular, is given by $\hat{\boldsymbol{V}}_{\boldsymbol{\theta}}(s) = \langle \x, {\boldsymbol{{\phi}}}_s'\rangle$. Throughout, we will make the standard assumption \cite{bhandari2018} that the columns of $\boldsymbol{\Phi}$ are independent and that the rows are normalized, i.e., $\|\boldsymbol{\phi}'_s\|_2^2\leq 1, \forall s \in \mathcal{S}.$

\textbf{Communication Model and \texttt{QFedTD} Algorithm.} Given the above setup, the goal of the server-agent system is to collectively estimate the model vector $\boldsymbol{\theta}^*$ corresponding to the best linear approximation of $\boldsymbol{V}_{\mu}$ in the span of $\boldsymbol{\Phi}$. To achieve this goal, we now describe a multi-agent variant of the classical \texttt{TD}(0) algorithm \cite{sutton1988learning}. All agents start out from a common initial state $s_0 \in \mathcal{S}$ with an initial estimate $\boldsymbol{\theta}_0 \in \mathbb{R}^{m}$. Subsequently, at each time-step $k \in \mathbb{N}$, a global model vector $\xk$ is broadcasted by the server to all agents. Each agent $i \in [N]$ then takes an action $a_{i,k}=\mu(s_{i,k})$, and observes the next state $s_{i,k+1} \sim \mathbb{P}_{\mu}(\cdot| s_{i,k})$ and instantaneous reward $r_{i,k}=R_{\mu}(s_{i,k})$; here, $s_{i,k}$ is the state of agent $i$ at time-step $k$. Using the model vector $\xk$ and the observation tuple $o_{i,k} = (s_{i,k}, r_{i,k}, s_{i,k+1})$, agent $i$ computes the following local TD update direction: 
\begin{equation*}
	\mathbf{g}_{i,k}(\xk, o_{i,k}) = (r_{i,k} + \gamma\langle \boldsymbol{\phi}'_{s_{i,k+1}}, \xk \rangle - \langle \boldsymbol{\phi}'_{s_{i,k}}, \xk \rangle)\boldsymbol{\phi}_{s_{i,k}}'. 
 \label{eqn:localTD}
\end{equation*}
We will often use $\mathbf{g}_{i,k}(\xk)$ as a shorthand for $\mathbf{g}_{i,k}(\xk, o_{i,k})$. Note that although all agents play the same policy $\mu$, and interact with the same MDP, the realizations of the local observation sequences $\{o_{i,k}\}$ can differ across agents. We  
 assume that these observation sequences are \textit{statistically 
 independent} across agents.\footnote{Notice that for each agent $i$, the observations over time are, however, correlated since they are all part of a single Markov chain.} Intuitively, based on this independence property, one can expect that exchanging agents' local TD update directions should help reduce the variance in the estimate of $\boldsymbol{\theta}^*$. This is precisely where the communication-induced challenges we describe below play a role. 

\textit{Channel Effects.} We model two key aspects of realistic communication channels in large-scale FL settings: finite capacity (due to limited bandwidth) and erasures/packet drops. To account for the first issue, we will employ a simple unbiased quantizer which is a (potentially random) mapping $\mathcal{Q}: \mathbb{R}^m \rightarrow \mathbb{R}^m$ satisfying the following constraints \cite{beznosikov}. 

\begin{definition} \label{defn:quant} (\textbf{Unbiased Quantizer}) We say that a quantizer $\mathcal{Q}$ is unbiased if the following hold for all $\mathbf{x} \in \mathbb{R}^m$:  (i) $\mathbb{E}\left[\mathcal{Q}(\mathbf{x})\right]=\mathbf{x}$, and (ii) there exists some  constant $\zeta \geq 0$ such that $\mathbb{E}\left[\Vert \mathcal{Q}(\mathbf{x}) - \mathbf{x} \Vert^2_2 \right] \leq \zeta \Vert \mathbf{x} \Vert^2_2$, where the expectation is w.r.t. the randomness of the quantizer. 
\end{definition}

The constant $\zeta$ captures the amount of distortion introduced by the quantizer. Using \textit{any} quantizer that satisfies Definition \ref{defn:quant}, each agent $i$ computes an encoded version $\mathbf{h}_{i,k}(\xk) = \mathcal{Q}(\mathbf{g}_{i,k}(\xk))$ of $\mathbf{g}_{i,k}(\xk)$. Here, we assume that the randomness of the quantizer is independent across agents and also independent of the Markovian observation tuples. 

Next, to capture packet drops, we assume that the encoded TD directions are uploaded to the server over Bernoulli erasure channels. Specifically, the transmission of information from an agent $i$ to the server is over a channel whose statistics are governed by an i.i.d. random process $\{b_{i,k}\}$, where for each $k$, $b_{i,k}$ follows a Bernoulli fading distribution. To be more precise, $b_{i,k}=0$ with erasure probability $(1-p)$, and $b_{i,k}=1$ with probability $p$. The packet-dropping processes are assumed to be independent of all other sources of randomness in our model. 

We are now in a position to describe the global model-update rule at the server:
\begin{equation}
	\boxed{\xkp = \xk + \alpha\mathbf{v}_k; \hspace{2mm} \vk = \frac{1}{N}\sum_{i = 1}^{N}b_{i,k} \mathbf{h}_{i,k}(\xk),}
 \label{eqn:globalupdate}
\end{equation}
where $\alpha$ is a constant step-size/learning rate. We refer to the overall updating scheme described above as the Quantized Federated TD learning algorithm, or simply $\texttt{QFedTD}$. 

\textbf{Objective and Challenges.} The main goal of this paper is to provide a \textit{finite-time analysis} of \texttt{QFedTD}. This is non-trivial for several reasons. Even in the single-agent setting, providing a non-asymptotic analysis of \texttt{TD}(0) without any projection step is known to be quite challenging due to temporal correlations between the Markov samples. To analyze \texttt{QFedTD}, \textit{we need to contend with three distinct sources of randomness}: (i) randomness due to the temporally correlated Markov samples $\{o_{i,k}\}_{i\in [N]}$; (ii) randomness due to the quantization step; and (iii) randomness due to the Bernoulli packet dropping processes $\{b_{i,k}\}_{i\in [N]}$. Each of these sources of randomness influence the evolution of the parameter vector $\xk$. Furthermore, unlike a single-agent setting, our goal is to establish a ``linear speedup" w.r.t. the number of agents under the different sources of randomness above. This necessitates a very careful analysis that we provide in Section~\ref{sec:Analysis}.

\begin{remark}
We note here that both the quantization mechanism and the channel model studied in this paper are quite simple. We have chosen to stick to these models primarily because the focus of our paper is on establishing the linear speedup effect under Markovian sampling. That said, we conjecture that the analysis in Section~~\ref{sec:Analysis} can potentially be extended to cover more involved encoding schemes (e.g., the use of error-feedback \cite{mitra2023temporal}), and more realistic channels with noise, interference, and non-stationary behavior. We reserve these questions for future work.
\end{remark}
\newpage
\section{Main result}
\label{sec:main_result}
In this section, we state and discuss our main result pertaining to the non-asymptotic performance of \texttt{QFedTD}. First, however, we need some technical preparation. As is standard, we assume that the rewards are uniformly bounded, i.e., $\exists \bar{r} > 0$ such that $R_{\mu}(s) \leq \bar{r}, \forall s \in \mathcal{S}.$ This ensures that the value function in \eqref{eqn:Value_func} is well-defined. Next, we make a standard assumption that plays a key role in the finite-time analysis of TD learning algorithms \cite{bhandari2018,srikant2019finite,tsitsiklisroy}. 

\begin{assumption}
The Markov chain induced by the policy $\mu$ is aperiodic and irreducible.
\label{assump:mixing}
\end{assumption}

An immediate consequence of the above assumption is that the Markov chain induced by $\mu$ admits a unique stationary distribution $\pi$ \cite{levin2017markov}. Let $\boldsymbol{\Sigma} = \boldsymbol{\Phi}^\top\mathbf{D}\boldsymbol{\Phi}$, where $\mathbf{D}$ is a diagonal matrix with entries given by the elements of the stationary distribution $\pi$. Since $\boldsymbol{\Phi}$ is assumed to be full column rank, $\boldsymbol{\Sigma}$ is full rank with a strictly positive smallest eigenvalue $\omega <1$; $\omega$ will later show up in our convergence bounds. Next, we define the steady-state local TD update direction as follows:
\begin{equation}
	\barg(\x) \triangleq  \mathbb{E}_{s_{i,k} \sim \pi, s_{i,k+1} \sim \mathbb{P}_{\mu}(\cdot| s_{i,k})}\left[\mathbf{g}_{i,k}(\x, o_{i,k})\right], \forall \x \in \mathbb{R}^m. 
\end{equation}

Essentially, the \textit{deterministic} recursion $\xkp = \xk + \alpha\barg(\xk)$ captures the limiting behavior of the \texttt{TD}(0) update rule. In \cite{bhandari2018}, it was shown that the iterates generated by this recursion converge exponentially fast to $\boldsymbol{\theta}^*$, where $\xs$ is the unique solution of the projected Bellman equation $\Pi_{\mathbf{D}}\mathcal{T}_{\mu}(\boldsymbol{\Phi}\boldsymbol{\xs})= \boldsymbol{\Phi}\boldsymbol{\xs}$. Here, $\Pi_{\mathbf{D}}(\cdot)$ is the projection operator onto the subspace spanned by $\{\boldsymbol{\phi}_\ell\}_{\ell \in [m]}$ with respect to the inner product $\langle \cdot, \cdot \rangle_{\mathbf{D}},$ and $\mathcal{T}_\mu:\mathbb{R}^{n} \rightarrow \mathbb{R}^{n}$ is the policy-specific Bellman operator \cite{tsitsiklisroy}. We now define the notion of mixing time $\tau_{\epsilon}$ that will play a crucial role in our analysis.

\begin{definition} \label{def:mix} 
Let $\tau_{\epsilon}$ be the minimum time such that the following holds: $$\Vert \mathbb{E}\left[\mathbf{g}_{i,k}(\boldsymbol{\theta}, o_{i,k})|o_{i,0}\right]-\barg(\boldsymbol{\theta})\Vert \leq \epsilon\left(\Vert \boldsymbol{\theta} \Vert +1 \right), {\forall k \geq \tau_{\epsilon},} \forall \boldsymbol{\theta} \in \mathbb{R}^m, \forall i \in [{N}], {\forall o_{i, 0}}.\footnote{Unless otherwise specified, we use $\Vert \cdot \Vert$ to denote the Euclidean norm.}$$
\end{definition}

{ Assumption \ref{assump:mixing} implies that the Markov chain induced by $\mu$ {mixes at a geometric rate \cite{levin2017markov}}, i.e., the total variation distance between $\mathbb{P}\left(s_{i,k}=\cdot|s_{i,0}=s\right)$ and the stationary distribution $\pi$ decays exponentially fast $\forall k \geq 0, \forall i \in [N], \forall s \in \mathcal{S}$. This immediately implies the existence of some $K \geq 1$ such that $\tau_{\epsilon}$ in Definition~\ref{def:mix} satisfies $\tau_{\epsilon} \leq K \log(\frac{1}{\epsilon})$~\cite{chenQ}. Loosely speaking, this means that for a fixed $\boldsymbol{\theta}$, if we want the noisy TD update direction to be $\epsilon$-close (relative to $\boldsymbol{\theta}$) to the steady-state TD direction (where both these directions are evaluated at $\boldsymbol{\theta}$), then the amount of time we need to wait for this to happen scales logarithmically in the precision $\epsilon.$ } For our purpose, the precision we will require is $\epsilon = \alpha^q$, where $q$ is an integer satisfying $q\geq 2$. Unlike the centralized case where $q=1$ suffices \cite{bhandari2018, srikant2019finite}, to establish the linear speedup property, we will require $q\geq2$. Henceforth, we will drop the subscript of $\epsilon=\alpha^q$ in $\tau_{\epsilon}$ and simply refer to $\tau$ as the mixing time. Let us define by $\sigma \triangleq \max \{1, \bar{r}, \Vert \xs \Vert \}$ the ``variance" of the observation model for our problem. Finally, let $\zeta' \triangleq \text{max}\{1, \zeta\}$, where $\zeta$ is {as in Definition \ref{defn:quant}}, and $\dxd \triangleq \dx$. We are now in a position to state the main result of this paper. 

\begin{theorem} \label{th:main}
	Consider the update rule of \texttt{QFedTD} in (\ref{eqn:globalupdate}). There exist universal constants {$C_0, C_2, C_3 \geq 1$}, such that with $\alpha \leq \frac{\omega(1-\gamma)}{C_0\tau\zeta'}$, the following holds for $T \geq 2\tau$: 
\begin{equation}
\label{eqn:main_bound}
\E{\delta_T^2} \leq (1 - \alpha\omega(1-\gamma)p)^{T}C_1+ \frac{\tau \sigma^2}{\omega (1-\gamma)} \left(\frac{C_2\alpha \zeta'}{N}+C_3 \alpha^3 \right),
\end{equation}
where $C_1 = 4\delta_0^2 + 2p\sigma^2$. 
\end{theorem}
{ 
\textbf{Discussion:} There are several important takeaways from Theorem \ref{th:main}. From \eqref{eqn:main_bound}, we first note that \texttt{QFedTD} guarantees linear convergence (in expectation) to a ball around $\xs$ whose radius depends on the variance $\sigma^2$ of the noise model. While the linear convergence rate gets slackened by the probability of successful transmission $p$, the ``variance term", namely the second term in \eqref{eqn:main_bound}, gets inflated by the quantization parameter $\zeta$. Both of these channel effects are consistent with what one typically observes for analogous settings in FL \cite{FedPAQ}. Next, compared to the centralized setting \cite[Theorem 7]{srikant2019finite}, the variance term in \eqref{eqn:main_bound} gets scaled down by a factor of $N$, up to a higher-order $O(\alpha^3)$ term that can be dominated by the $(\alpha/N)$ term for small enough $\alpha$. Before we make this point explicit, it is instructive to note that our variance bound exhibits a tighter dependence on the mixing time $\tau$ relative 
to \cite{khodadadian2022federated} and \cite{han}, where similar bounds are established. In particular, while this dependence is $O(\tau)$ for us, it is $O(\tau^2)$ in \cite[Theorem 4.1]{khodadadian2022federated} and in \cite[Theorem 4]{han}. Notably, the $O(\tau)$ dependence that we establish is consistent with results on centralized TD learning~\cite{bhandari2018, srikant2019finite}, and is in fact the optimal dependence on $\tau$ under  Markovian data \cite{nagaraj}. We have the following immediate corollary of Theorem~\ref{th:main}.

\begin{corollary}
\label{corr:linspeed}
    Consider the update rule of \texttt{QFedTD} in~\eqref{eqn:globalupdate}. Let the step-size $\alpha$ and the number of iterations $T$ be chosen to satisfy:
    \begin{equation}
    \alpha = \frac{\log NT}{ \omega (1-\gamma) pT}, \hspace{2mm} \textrm{and} \hspace{2mm} T \geq \frac{ 2 C_0 N \tau \zeta' \log NT}{  \omega^2 (1-\gamma)^2 p},
\end{equation}
where $C_0$ is as in Theorem~\ref{th:main}. We then have the following bound: 
\begin{equation}
    \mathbb{E}[{\delta_T^2}] \leq O\left( {\tb \left({ \frac{\zeta'}{p}}\right) }  \frac{ \max\{\delta^2_0, \sigma^2\} \tau \log (NT)}{\omega^2 (1-\gamma)^2 {\tr {NT}}} \right).
\label{eqn: simplified bound}   
\end{equation}
\end{corollary}

To appreciate the above result, let us compare it to the result for the single agent TD setting in \cite{bhandari2018}. Under Markovian sampling, part (c) of Theorem 3 in \cite{bhandari2018} establishes that the mean-square error for single-agent TD decays at the following rate:
$$ O \left( \frac{G^2 \tau \log(T)} {\omega^2 (1-\gamma)^2 T}                  \right),$$
where $G$, as defined in \cite{bhandari2018}, captures the effect of both the projection radius (in \cite{bhandari2018}, the authors consider a projected version of TD learning) and the noise variance.\footnote{Part (c) of Theorem 3 in \cite{bhandari2018} provides a bound on the error in the value function, and not the iterates (like we do). The bound on the iterates that we report above is derived from the bound on the value functions in Appendix A.2 of \cite{bhandari2018}, where the authors provide a proof of Theorem 3.} The term $G^2$ can be viewed as the analog of $\max\{\delta^2_0, \sigma^2\}$ in our bound. Comparing the above bound with that in Eq.~\eqref{eqn: simplified bound}, we make two immediate observations. (i) The term $T$ in the centralized bound gets replaced by $NT$ in our bound. This is precisely what we wanted since in our setting, each agent has access to $T$ samples, yielding a total of $NT$ samples. Essentially, this goes on to show that our algorithm is \textit{sample-efficient} in that it makes use of all the samples from all the agents and achieves a linear speedup w.r.t. the number of agents. Second, the effect of channel effects is succinctly captured by the term in blue in Eq.~\eqref{eqn: simplified bound}. This term essentially inflates the variance $\max\{\delta^2_0, \sigma^2\}$ of our noise model. When the number of agents $N=1$, the probability of successful transmission $p=1$, and there is no quantization effect (i.e., $\zeta'=1$), our bound exactly recovers the bound in the centralized setting (even up to log factors). As far as we are aware, our work is the first to establish such a tight result in multi-agent/federated reinforcement learning under Markovian sampling and communication constraints.}

\newpage
\section{Proof of the Main Result}
\label{sec:Analysis}
In this section, we will prove Theorem \ref{th:main}. We start by introducing some definitions to lighten the notation, and by recalling some basic results from prior work. Let us define: 
\begin{equation}\label{eq:dxddifftau}
\begin{aligned}
    \muiktau(\x) & \triangleq \|\E{\mathbf{g}_{i,k}(\x, \okn)|o_{i, k-\tau}} - \barg(\x)\|,\ k\geq \tau, \\
    \delta_{k, \tau} & \triangleq \|\xk - \x_{k-\tau}\|,\ k\geq \tau. 
\end{aligned}
\end{equation}
For our analysis, we will need the following result from \cite{bhandari2018}. 
\begin{lemma}\label{lemma:bhand}
	The following holds $\forall \boldsymbol{\theta} \in \mathbb{R}^m$: 
\begin{equation*}\label{eq:bhandlemmas}
\begin{aligned}
		\langle \xs - \x, \barg(\x) \rangle \geq \omega(1-\gamma)\|\xs -\x\|^2.
\end{aligned}	
\end{equation*}
\end{lemma}\vspace{0.3cm}
We will also use the fact that the random TD update directions and their steady-state versions are 2-Lipschitz \cite{bhandari2018}, i.e., $\forall i \in [N], \forall k \in \mathbb{N}$, and $\forall \x, \x' \in \mathbb{R}^m$, we have: 
\begin{equation}\label{eq:Lipschitz}
\begin{aligned}
   \max\{\|\gkn(\x) - \gkn(\x')\|, \|\bar{\mathbf{g}}(\x) - \bar{\mathbf{g}}(\x')\| \} \leq 2\|\x - \x'\|.
\end{aligned}
\end{equation}
Finally, we will use the following bound from \cite{srikant2019finite}: 
\begin{equation}\label{eq:boundGradNorm}
    \|\mathbf{g}_{i,k}(\x, \okn)\|\leq 2\|\x\| + 2\bar{r}, \forall i \in [N], \forall k \in \mathbb{N},\forall \x\in \mathbb{R}^m.
\end{equation}
Equipped with the above basic results, we now provide an outline of our proof before delving into the technical details.

{\textbf{Outline of the proof}. We start by defining:
\begin{equation}
\begin{aligned}
	\barg_N(\xk) &\triangleq \frac{1}{N}\sum_{i = 1}^{N}\bkn\barg(\xk), \hspace{2mm} \textrm{and} \\
 \psi_k &\triangleq \langle \vk - \barg_N(\xk),\xk - \xs \rangle. 
\end{aligned}
\end{equation}
Since for all $i\in [N]$, $b_{i,k}$ is independent of $\xk$, we have $\E{\langle \barg_N(\xk),\xk - \xs \rangle} = p\E{\langle \barg(\xk),\xk - \xs \rangle}$. Thus, recalling that $\dxd \triangleq \dx$, and using \eqref{eqn:globalupdate}, we obtain
\begin{equation}
	\begin{aligned}
		\E{\dxpd} &= \E{\dxd} -2\alpha\E{\langle \xs - \xk, \vk \rangle} + \alpha^2\E{\|\vk\|^2}
		\\&= \E{\dxd} - 2\alpha p\E{\langle \xs - \xk, \barg(\xk) \rangle} \\&\hspace{2mm}+ 2\alpha\E{\psi_k} + \alpha^2\E{\|\vk\|^2}.
	\end{aligned}
 \label{eq:recursionMain}
\end{equation}}{The main technical burden in proving Theorem \ref{th:main} is in bounding $\E{\|\vk\|^2}$ and $\E{\psi_k}$ in the above recursion. Following the centralized analysis in \cite{bhandari2018, srikant2019finite}, one can easily bound $\E{\|\vk\|^2}$ using \eqref{eq:boundGradNorm}. However, this approach will fall short of yielding the desired linear speedup property. Hence, to bound $\E{\|\vk\|^2}$, we need a much finer analysis, one that we provide in Lemma~\ref{lemma:claim_1}. Leveraging Lemma~\ref{lemma:claim_1}, we then establish an intermediate result in Lemma~\ref{lemma:claim_2} that bounds $\E{\|\xk - \x_{k-\tau}\|}$. This result, in turn, helps us bound $\E{\psi_k}$ in Lemma~\ref{lemma:claim_3}. We now proceed to flesh out these steps. In what follows, $\tau = \tau_{\epsilon}$ with $\epsilon = \alpha^q$, $q\geq 2$. 
\begin{lemma} (\textbf{Key Technical Result}) \label{lemma:claim_1} For $k\geq\tau$, we have
\begin{equation}
 \begin{aligned}
		\E{\|\vk\|^2} \leq 60\zeta'p\E{\dxd} + 12\sigma^2p\left(10\frac{\zeta'}{N} + \alpha^{2q}\right). 
 \end{aligned}	
 \end{equation}
\begin{proof}
Note that $\|\vk\|^2 \leq \frac{3}{N^2}(T_1+T_2+T_3)$, with
	\begin{equation}\label{eq:T123_Cl1}
	\begin{aligned}
				T_1 &= \Vert \sum_{i = 1}^{N}\bkn\gkn(\xs) \Vert^2, \\ 
				T_2 &= \Vert \sum_{i = 1}^{N}\bkn(\gkn(\x_k) - \gkn(\xs)) \Vert^2, \hspace{2mm} \textrm{and} \\
				T_3 &= \Vert\sum_{i = 1}^{N}\bkn(\gkn(\x_k) - \hkn(\xk))\Vert^2. 
	\end{aligned}
	\end{equation}
We now proceed to bound $T_1-T_3$. To that end, we first write $T_1$ as 
\eqal{}{
T_1 &= T_{11}+ T_{12},\hspace{0.1cm} \textrm{with}\\
T_{11} &= \sum_{i = 1}^{N}\bkn^2\|\gkn(\xs)\|^2, \hspace{0.1cm}\textrm{and}\\  T_{12} &= \sum_{\substack{i,j=1\\i\neq j}}^{N}\bkn\bkm\langle\gkn(\xs), \gkm(\xs) \rangle.
} 
Now using \eqref{eq:boundGradNorm}, we obtain  $T_{11} \leq 8(\|\xs\|^2 + \bar{r}^2)\sum_{i = 1}^{N}\bkn^2
$. Recalling that $\sigma \triangleq \max \{1, \bar{r}, \Vert \xs \Vert \}$, we then have $\E{T_{11}} \leq 16\sigma^2\E{\sum_{i = 1}^{N}\bkn^2}= 16\sigma^2Np$. Next, to bound the cross-terms in $T_{12}$, we will exploit the mixing property in Definition~\ref{def:mix}. To that end, we note that since (i) $\barg{(\xs)} = \mathbf{0}$ \cite{bhandari2018}, (ii) the packet-dropping processes are independent of the Markovian tuples, and (iii) $\gkn(\xs)$ and $\gkm(\xs)$ are independent for $i\neq j$, 
\eqal{}{
\E{T_{12}} = \sum_{\substack{i,j = 1\\i\neq j}}^{N}\E{\bkn\bkm}\langle \E{\E{\gkn(\xs)|\okntau}-\barg{(\xs)}},\E{\E{\gkm(\xs)|\okmtau}-\barg{(\xs)}} \rangle. \nonumber
}
Using the Cauchy-Schwarz inequality followed by Jensen's inequality, we can further bound the above inner-product via $\E{\muiktau(\xs)} \times \E{\mujktau(\xs)} \leq 4 \sigma^2 \alpha^{2q}$. For the last inequality, we used the mixing property by noting that $k \geq \tau$. Specifically, appealing to Definition~\ref{def:mix}, and recalling that $\sigma \triangleq \max \{1, \bar{r}, \Vert \xs \Vert \}$, we have
$$ \muiktau(\xs) \leq \alpha^q(\Vert \xs \Vert +1) \leq 2\sigma \alpha^q.$$
Clearly, the same bound also applies to $\mujktau(\xs)$ via an identical reasoning. Combining this analysis with the fact that $\E{\bkn\bkm} = \E{\bkn} \E{\bkm} = p^2$, we obtain that $\E{T_{12}}\leq 4N^2p^2\sigma^2\alpha^{2q}.$  Combining the bounds for $\E{T_{11}}$ and $\E{T_{12}}$ thus yields: 
\eqal{}{
\E{T_1} \leq 16\sigma^2Np + 4N^2p^2\sigma^2\alpha^{2q}.
}
Now, using \eqref{eq:Lipschitz}, we see that
\begin{equation}
	\begin{aligned}
		\E{T_2} &\leq N\sum_{i = 1}^{N}\E{\bkn^2\|\gkn(\xk) - \gkn(\xs)\|^2}\\&\leq 4N\E{\dxd}\sum_{i = 1}^{N}\E{\bkn^2} = 4pN^2\E{\dxd}.
	\end{aligned}
\end{equation}
Defining $\boldsymbol{\lambda}_{i, k}(\xk) \triangleq \hkn(\xk) - \gkn(\xk)$, we now turn to bounding $T_3$ by writing it as  
\begin{equation}
	\begin{aligned}
        T_3 &= T_{31}+T_{32}, \hspace{0.1cm}\textrm{with}\\
		T_{31} &= \sum_{i = 1}^{N} \bkn^2\|\boldsymbol{\lambda}_{i, k}(\xk)\|^2, \hspace{2mm} \textrm{and} \\
		T_{32} &= \sum_{\substack{i,j\\i\neq j}}^{N}\bkn\bkm\langle \boldsymbol{\lambda}_{i, k}(\xk),\boldsymbol{\lambda}_{j, k}(\xk) \rangle.
	\end{aligned}
\end{equation}

We now proceed to bound $\E{T_{31}}$ and $\E{T_{32}}$ as follows: 
\begin{equation*}
\begin{aligned}
        \E{T_{31}} & = \sum_{i = 1}^{N}\E{\bkn^2}\E{\E{\boldsymbol{\|\lambda}_{i,k}(\xk)\|^2|\okn, \xk}}\\
& \overset{(a)} \leq \sum_{i = 1}^{N}p\zeta\E{\|\gkn(\xk)\|^2}\\& \overset{(b)}\leq 8Np\zeta(\E{\|\xk\|^2} + \sigma^2)\\
&\leq 16Np\zeta\E{\|\xk-\xs\|^2} + 24Np\zeta\sigma^2,
\end{aligned}
\end{equation*}
where $(a)$ follows from the variance bound of the quantizer map $\mathcal{Q}(\cdot)$, and $(b)$ follows from \eqref{eq:boundGradNorm}. Next, observe that: 
\eqal{}{
\E{T_{32}} = p^2\sum_{\substack{i,j = 1\\i\neq j}}^{N}\E{\E{\langle \boldsymbol{\lambda}_{i, k}(\xk),\boldsymbol{\lambda}_{j, k}(\xk) \rangle|o_{i,k}, o_{j,k}, \xk}}. \nonumber
}
Using the fact that the randomness of the quantization map is independent across agents, and the unbiasedness of $\mathcal{Q}(\cdot)$, we conclude that $\E{T_{32}}=0$. Combining the bounds on $\E{T_{1}}$, $\E{T_{2}}$, and $\E{T_{3}}$ above yields the desired result. 
\end{proof}
\end{lemma}

\begin{remark} As the rest of our analysis will reveal, Lemma~\ref{lemma:claim_1} is really the key technical result that will help us establish the desired linear speedup effect under Markovian sampling. One important takeaway from the proof of this result is that we do not need to exploit the fact that the TD update direction is an affine function of the parameter $\xk$. As such, Lemma~\ref{lemma:claim_1} should essentially be applicable (with potentially minor modifications) to more general stochastic approximation schemes where the operator under consideration satisfies basic smoothness properties.
\end{remark}

Later in the analysis, we will once again need to invoke a mixing time argument by conditioning on $\xktau$. This will give rise to the {$\delta_{k, \tau} = \|\xk - \x_{k-\tau}\|$} term that we proceed to bound below by leveraging Lemma~\ref{lemma:claim_1}. 

\begin{lemma}\label{lemma:claim_2}
	Let $\alpha\leq \frac{1}{484\tau\zeta'}$ and $k \geq 2\tau$. Then, we have
	\begin{equation}
		 \E{\delta_{k, \tau}^2} \leq 480\alpha^2\tau^2p\zeta'\E{\dxd}+ \alpha^2\tau^2p\sigma^2\left(\frac{360\zeta'}{N} + 4\alpha^q \right). 
\nonumber
	\end{equation}
\begin{proof}
We start with a bound on $\dxpd$:
\begin{equation}\label{eq:boundDeltaLess}
		\begin{aligned}
			\dxpd &= \dxd -2\alpha\langle \vk, \xs - \xk \rangle + \alpha^2\|\vk\|^2\\
             &\overset{(a)} \leq \dxd + 2\alpha \Vert \vk \Vert   \delta_k  + \alpha^2\|\vk\|^2\\
            & \overset{(b)} \leq (1 + \alpha)\dxd + (\alpha + \alpha^2)\|\vk\|^2\\
            & \overset{(c)} \leq(1 + \alpha)\dxd + 2\alpha\|\vk\|^2.
		\end{aligned}
	\end{equation}
In the above steps, (a) follows from the Cauchy-Schwarz inequality. For (b), we note that given any two positive numbers $x$ and $y$, it holds that
$$ xy \leq \frac{1}{2} x^2 + \frac{1}{2} y^2.$$
For (c), we simply used the fact that since $\alpha \in (0,1)$, it holds that $\alpha^2 \leq \alpha$. Hence, $\alpha + \alpha^2 \leq 2 \alpha.$ Using Lemma \ref{lemma:claim_1} and the fact that $p < 1$, we obtain 
\begin{equation}
\begin{aligned}
	\E{\dxpd} \leq (1+121\alpha\zeta')\E{\dxd}+\underbrace{24\alpha p\sigma^2{\left(\frac{10\zeta'}{N} + \alpha^{2q}\right)}}_{B}.
 \nonumber 
\end{aligned}
\end{equation}
Iterating this inequality, we get for any $k-\tau\leq k' \leq k$, 
\begin{equation}
	\begin{aligned}
		\E{\delta^2_{k'}} &
  \leq (1 + 121\alpha\zeta')^{\tau}\E{\dxdtau} + B \sum_{\ell = 0}^{\tau-1}(1 + 121\alpha\zeta')^{\ell}.
	\end{aligned}
\label{eqn:sum}
\end{equation}
Now using $(1+x) \leq e^x, \forall x \in \mathbb{R}$, observe that $(1 + 121\alpha\zeta')^{\ell} \leq (1 + 121\alpha\zeta')^{\tau} \leq e^{0.25} \leq 2$, for $\alpha\leq 1/(484\tau \zeta').$ Thus, 
$\sum_{\ell = 0}^{\tau-1}(1 + 121\alpha \zeta')^{\ell} \leq 2\tau$. Plugging this bound in \eqref{eqn:sum}, we obtain 
\begin{equation}\label{eq:iterated}
	\E{\delta^2_{k'}}\leq 2\E{\dxdtau} + 2\tau B.
\end{equation}
Next, observe that $$
	\dxddifftau \leq \tau\sum_{\ell = k-\tau}^{k-1}\|\x_{\ell+1} - \x_{\ell}\|^2 = \tau\alpha^2\sum_{\ell = k-\tau}^{k-1}\|\mathbf{v}_{\ell}\|^2.$$
Since $k \geq 2 \tau$, we have $\ell \geq \tau$. Hence, we can invoke Lemma \ref{lemma:claim_1} to bound $\E{\|\mathbf{v}_{\ell}\|^2}$. This yields
\begin{equation}
	\E{\dxddifftau}\leq \alpha^2\tau\sum_{\ell = k-\tau}^{k-1}60\zeta' p\E{\delta^2_{\ell}} + 0.5 \alpha \tau^2 B.
\end{equation}
Using (\ref{eq:iterated}) to bound $\E{\delta^2_{\ell}}$ above, we further obtain 
 \eqalNo{}{
\E{\dxddifftau} \leq \alpha^2\tau\sum_{\ell = k-\tau}^{k-1}120\zeta' p\left(\E{\dxdtau}+\tau B\right)+\frac{1}{2} \alpha \tau^2 B.
}
Simplifying using $\alpha \leq 1/484\zeta'\tau$, $p<1$, and $q\geq2$ yields
\begin{equation}
	\E{\dxddifftau}\leq 120\alpha^2\tau^2p\zeta'\E{\dxdtau} + \alpha^2\tau^2\sigma^2p\left(\frac{180\zeta'}{N} + 2\alpha^q\right).
 \nonumber 
\end{equation}
Using $\delta^2_{k-\tau} \leq 2 \delta^2_{k}+2\delta^2_{k,\tau}$ and  $240\alpha^2\tau^2\zeta'\leq 1/2$ to simplify the above inequality, we arrive at the desired result. 
\end{proof}
\end{lemma}

Our next result is the final ingredient needed to prove Theorem \ref{th:main}. 

\begin{lemma}\label{lemma:claim_3} Define $$\mathbf{g}_N(\xk) \triangleq \frac{1}{N}\sum_{i = 1}^{N}\bkn\gkn(\xk),$$ and let $\alpha\leq 1/(484\zeta' \tau)$ and $k \geq 2\tau$. We have
\begin{equation}
\begin{aligned}
	\E{\psi_k}\leq \alpha\tau p\left(3191\zeta'\E{\dxd} + \sigma^2\left(\frac{2461\zeta'}{N} + {30}\alpha^q\right)\right). 
 \nonumber
\end{aligned}
\end{equation}
\begin{proof}
We can write $\psi_k = T_1 + T_2 + T_3 + T_4 + T_5$, with 
\eqal{}{
T_1 &= \langle \xk - \x_{k-\tau}, \bfg_N(\x_{k}) - \barg_N(\xk) \rangle,  \\
T_2 &= \langle \x_{k-\tau} - \xs, \bfg_N(\x_{k-\tau}) - \barg_N(\x_{k-\tau}) \rangle,  \\
T_3 &= \langle \x_{k-\tau} - \xs, \bfg_N(\x_{k}) - \bfg_N(\x_{k-\tau}) \rangle,  \\
T_4 &= \langle \x_{k-\tau} - \xs, \barg_N(\x_{k-\tau}) - \barg_N(\xk) \rangle, \\
T_5 &= \langle \x_{k} - \xs,  \vk - \bfg_N(\x_{k}) \rangle.  
}
To bound $T_1$, observe the following inequalities:
\eqal{eq:T_1_1}{
    T_1 &= \langle \xk - \x_{k-\tau}, \bfg_N(\x_{k}) - \barg_N(\xk) \rangle
    \\&
    \overset{(a)}\leq \|\xk - \x_{k-\tau}\|\|\bfg_N(\x_{k}) - \barg_N(\xk)\|
    \\&
    \overset{(b)}\leq \frac{1}{2\alpha\tau}\|\xk - \x_{k-\tau}\|^2 + \frac{\alpha\tau}{2}\|\bfg_N(\x_{k}) - \barg_N(\xk)\|^2
    \\&
    \overset{(c)}\leq \underbrace{\frac{1}{2\alpha\tau}\|\xk - \x_{k-\tau}\|^2}_{S_1}+\underbrace{\alpha\tau\|\bfg_N(\x_{k})\|^2}_{S_2} + \underbrace{\alpha\tau\|\barg_N(\xk) - \barg_N(\xs)\|^2}_{S_3}.
}
In the above steps, (a) follows from the Cauchy-Schwarz inequality. For (b), we used the fact that given any two positive numbers $x$ and $y$, the following holds for any $\eta >0$:
$$ x y \leq \frac{1}{2\eta} x^2 + \frac{\eta}{2} y^2.$$
We used the above inequality with $\eta=\alpha \tau$ to arrive at (b). Finally, for (c), we used the fact that $\bar{\bfg}(\xs)=0$; hence, $\bar{\bfg}_N(\xs)=0$. We now proceed to bound the expectations of each of the terms $S_1 - S_3$, starting with $S_3$. Note that using \eqref{eq:Lipschitz}, i.e., the Lipschitz property of the TD update directions, we get: 
\eqalNo{}{
\|\barg_N(\xk) - \barg_N(\xs)\|^2 &\leq \Vert \sumN b_{i, k}(\barg_N(\xk) - \barg_N(\xs))\Vert^2
\\&
\leq \frac{1}{N}\sum_{i = 1}^{N}\bkn^2\|\barg_N(\xk) - \barg_N(\xs)\|^2
\\&
\leq \frac{4}{N}\sum_{i = 1}^{N}\bkn^2\|\xk  - \xs\|^2.
}

Taking expectations on each side of the above inequality then yields:
\eqal{eq:T_1_2}{
\E{\alpha\tau\|\barg_N(\xk) - \barg_N(\xs)\|^2} \leq 4\alpha\tau p\E{\|\xk  - \xs\|^2}. 
}

In arriving at the above inequality, we used the following facts: (i) the randomness in $\xk$ depends on all the sources of randomness in our model up to time $k-1$; (ii) the Bernoulli packet-drop random variables $\{b_{i,k}\}_{i \in [N]}$ are independent of all the sources of randomness up to time $k-1$. Hence, for each $i\in [N]$, $\E{ b^2_{i,k} \|\xk  - \xs\|^2} = \E{ b^2_{i,k} } \E{ \|\xk  - \xs\|^2} = p \E{ \|\xk  - \xs\|^2}.$

Next, to bound $\E{S_1}$, note that $\E{\frac{1}{2\alpha\tau}\|\xk - \x_{k-\tau}\|^2}$ can be directly bounded using Lemma~\ref{lemma:claim_2} in the following way:
\eqal{eq:T_1_3}{
\frac{1}{2\alpha\tau}\E{\|\xk - \x_{k-\tau}\|^2} \leq 240\alpha \tau p\zeta'\E{\|\xk - \xs\|^2}+ \alpha \tau p\sigma^2\left(\frac{180\zeta'}{N} + 2\alpha^q \right).
}

Finally, the only term that remains to be bounded is $\E{\|\bfg_N(\x_{k})\|^2}$. Note that we can write: 
\eqal{eq:T_1_4}{
\|\bfg_N(\x_{k})\|^2 &\leq \frac{2}{N^2}(T_1' + T_2') \hspace{2mm} \textrm{with}\\
T_1' &= \Vert \sum_{i = 1}^{N}\bkn\gkn(\xs) \Vert^2, \hspace{2mm} \textrm{and}\\ 
T_2' &= \Vert \sum_{i = 1}^{N}\bkn(\gkn(\x_k) - \gkn(\xs)) \Vert^2.  \hspace{2mm}
}
Observe that $T_1'$ and $T_2'$ above correspond exactly to the terms  $T_1$ and $T_2$ in the proof of Lemma~\ref{lemma:claim_1}. Thus, they can be bounded as follows: 
\eqal{eq:T_1_5}{
\E{T_1'} &\leq 16\sigma^2Np + 4N^2p^2\sigma^2\alpha^{2q}.\\
\E{T_2'} &\leq 4pN^2\E{\|\xk - \xs\|^2}. 
}
So, plugging \eqref{eq:T_1_2}, \eqref{eq:T_1_3}, \eqref{eq:T_1_4}, and the above bound into \eqref{eq:T_1_1}, we get the final bound on $\E{T_1}$ as follows:
\begin{equation*}
	\E{T_1} \leq 304\alpha\tau\zeta' p\E{\dxd} + \alpha\tau p\sigma^2\left(\frac{300\zeta'}{N} + 3\alpha^q\right).
\end{equation*}
Next we bound $\E{T_3}$ and $\E{T_4}$. Observe that:
\begin{equation*}
\begin{aligned}
\E{T_3} &=  \sumN\E{\bkn\langle \x_{k-\tau} - \xs, (\gkn(\xk) - \gkn(\x_{k-\tau})) \rangle}\\&
\leq p\E{\delta_{k-\tau}\sumN{\Vert \gkn(\xk) - \gkn(\x_{k-\tau})\Vert}}\\&
\overset{\eqref{eq:Lipschitz}} \leq 2p\E{\delta_{k-\tau}\delta_{k, \tau}}
\\&
\leq \myFrac{\alpha\tau p}{2}\E{\dxdtau} + \myFrac{2p}{\alpha\tau}\E{\dxddifftau}.
\end{aligned}
\end{equation*}
Using $\delta^2_{k-\tau} \leq 2 \delta^2_{k}+2\delta^2_{k,\tau}$ and Lemma \ref{lemma:claim_2}, we then obtain:
\begin{equation}
	\E{T_3}\leq 1441\alpha\tau p\zeta'\E{\dxd} + 6\alpha\tau p\sigma^2\left(\frac{180\zeta'}{N} + 2\alpha^q\right).
 \nonumber
\end{equation}
Using the same process, we can derive the exact same bound for $\E{T_4}$. We now bound $\E{T_2}$. For ease of notation, let us define $\mathcal{F}_{k, \tau} = (\{o_{i, k-\tau}\}_{i = 1}^N, \x_{k-\tau})$. Observe: 
\begin{equation}
	\begin{aligned}
		\E{T_2} &= \E{\E{T_2|\mathcal{F}_{k, \tau}}}
		\\&
  = \mathbb{E}[\langle \x_{k-\tau} - \xs, \frac{p}{N}\sum_{i = 1}^{N} (\E{\mathbf{g}_{i,k}(\x_{k-\tau}, \okn)|\mathcal{F}_{k, \tau}} - \barg(\xktau))\rangle]\\&
		\leq \E{\delta_{k - \tau} \frac{p}{N} \sum_{i = 1}^{N}\muiktau(\x_{k-\tau})}\\&
		\leq {p\alpha^q}\E{\delta_{k - \tau}(1 + \|\x_{k-\tau}\|)},
	\end{aligned}
 \nonumber
\end{equation}
where in the last step, we made use of the mixing property. Since $\alpha <1$, we have $\delta_{k-\tau}(\delta_{k-\tau} + 2\sigma) \leq \frac{\dxdtau}{\alpha} + 2\sigma\delta_{k-\tau} +\alpha\sigma^2 = \left(\frac{\delta_{k-\tau}}{\sqrt{\alpha}} + \sqrt{\alpha}\sigma\right)^2 \leq 2\left(\frac{\dxdtau}{\alpha} + \alpha\sigma^2\right)$. Using $q \geq 2$, we obtain: 
\begin{equation}
	\begin{aligned}
		\E{T_2} &\leq 2p\alpha^q\E{\frac{1}{\alpha}\dxdtau + \alpha\sigma^2}\\&
		\leq 2p\alpha\E{\dxdtau} + 2p\alpha^{q+1}\sigma^2.
	\end{aligned}
\end{equation}
Using $\delta^2_{k-\tau} \leq 2 \delta^2_{k}+2\delta^2_{k,\tau}$ and Lemma \ref{lemma:claim_2}, and then simplifying yields: 
\begin{equation}
	\E{T_2} \leq 5\alpha\tau p\zeta'\E{\dxd} + \alpha\tau p\sigma^2 \left(\frac{\zeta'}{N} + 3\alpha^q\right).
\end{equation}
Finally, to bound $T_5$, let $\mathcal{F}_k = \{\{\okn\}_{i = 1}^N, \xk\}$. We have
\eqal{}{
\E{T_5} = \E{\langle \x_{k} - \xs,  \underbrace{\E{\vk - \bfg_N(\x_{k})|\mathcal{F}_k}}_{T_{51}} \rangle}.
}
Note that 
$T_{51} = \frac{p}{N}\sum_{i = 1}^{N}\E{\hkn(\xk) - \gkn(\xk)|\mathcal{F}_k} = 0$, based on the unbiasedness of $\mathcal{Q}(\cdot)$. Thus, $\E{T_5} = 0$. Collecting the bounds on $T_1-T_5$ concludes the proof.
\end{proof}
\end{lemma}
With the help of the auxiliary lemmas provided above, we are now ready to prove our main result, i.e., Theorem \ref{th:main}. 

\textbf{Proof of Theorem \ref{th:main}}.
{Setting $\alpha \leq \frac{1}{484\zeta'\tau}$, we can apply the bounds in Lemmas \ref{lemma:bhand}, \ref{lemma:claim_1}, and  \ref{lemma:claim_3} to \eqref{eq:recursionMain}. This yields}: 
\begin{equation}
		\E{\dxpd} \leq \E{\dxd} -\alpha p(2(1-\gamma)\omega - 6446\alpha\tau\zeta')\E{\dxd} + 5162\frac{\alpha^2\tau p\sigma^2\zeta'}{N} + 61\alpha^{(2+q)}\tau p\sigma^2.
\end{equation}
For $\alpha \leq \frac{\omega(1-\gamma)}{C_0\tau\zeta'}$  with $C_0 = 6446$, we then obtain: 
\begin{equation}
	\E{\dxpd} \leq (1-\alpha\omega(1-\gamma)p)\E{\dxd}  +5162\frac{\alpha^2\tau p\sigma^2\zeta}{N} + 61\alpha^{(2+q)}\tau p\sigma^2.
\end{equation}
Iterating the last inequality, we have $\forall k \geq 2\tau$: 
\begin{equation}
	\E{\dxd} \leq \rho^{k-2\tau}\E{\delta^2_{2\tau}}+\frac{\tau \sigma^2}{\omega (1-\gamma)} \left(\frac{C_2\alpha \zeta'}{N}+C_3 \alpha^3 \right), 
 \nonumber
\end{equation}
where $\rho=(1 - \alpha\omega(1-\gamma)p)$, $C_2 = 5162$, $C_3 = 61$, and we set $q=2$. It only remains to show that with our choice of $\alpha$, $\E{\delta^2_{2\tau}}=O(\delta^2_0 + \sigma^2)$. This follows from some simple algebra and steps similar to those in the proof of Lemma~\ref{lemma:claim_2}. We provide these steps below for completeness. 
Note that, defining $T' = \|\sum_{i = 1}^N\bkn\gkn(\xk)\|^2$, and using (\ref{eq:boundGradNorm}),
\begin{equation*}
\begin{aligned}
    \E{T'}\leq8N^2p\E{2\dxd + 3\sigma^2}\leq N^2(16p\E{\dxd} + 24p\sigma^2).
\end{aligned}
\end{equation*}
Letting $T_3$ be as defined in (\ref{eq:T123_Cl1}), note that
\begin{equation*}\label{eq:preTauvksq}
    \E{\|\vk\|^2}\leq \frac{2}{N^2}\E{T'+T_3} \leq 64p\zeta'\E{\dxd} + 96p\zeta'\sigma^2.
\end{equation*}
Plugging this inequality into (\ref{eq:boundDeltaLess}) and iterating,
\begin{equation*}
    \begin{aligned}
    \E{\delta_{k}^2}\leq (1+129\alpha\zeta')^{k}\dxzd + 192\alpha p\sigma^2\sum_{j = 0}^{k-1}(1+129\alpha\zeta')^j.
    \end{aligned}
\end{equation*}
Using the same arguments used to arrive at \eqref{eq:iterated}, we have
\begin{equation}
\begin{aligned}
    \E{\delta_{2\tau}^2}&\leq (1 + 129\alpha\zeta')^{2\tau}\dxzd + {768}\alpha\tau p\zeta'\sigma^2\\&\leq
    2\delta_0^2 + p\sigma^2,
\end{aligned}
\end{equation}
where we used the fact that $\alpha\tau \leq \frac{\omega(1-\gamma)}{6446\zeta'}\leq \frac{1}{{1032}\zeta'}$. Given that $\alpha\tau \leq \frac{1}{C_0}\leq \frac{1}{4}$, from Bernoulli's inequality, we have that $(1-\alpha)^{2\tau}\geq 1 - 2\alpha\tau \geq \frac{1}{2}$. Thus, observe that  $(1 - \alpha\omega(1-\gamma)p)^{-2\tau}\leq (1-\alpha)^{-2\tau}\leq 2$. This concludes the proof. 
\hspace*{\fill}$\square$
\newline
\newline
We now provide the proof of Corollary~\ref{corr:linspeed}.

\textbf{Proof of Corollary~\ref{corr:linspeed}}. We first recall the main result of Theorem~\ref{th:main}, i.e., the following bound:
\begin{equation}
\label{eqn:main_boundRec}
\E{\delta_T^2} \leq \underbrace{(1 - \alpha\omega(1-\gamma)p)^{T}C_1}_{T_1} + \underbrace{\frac{C_2\alpha \zeta'\tau \sigma^2}{\omega(1-\gamma)N}}_{T_2} +\underbrace{\frac{C_3\alpha^3\tau\sigma^2}{\omega (1-\gamma)}}_{T_3}.
\end{equation}
Let us also recall the choice of step-size $\alpha$ and number of iterations $T$ from Corollary~\ref{corr:linspeed}:
\begin{equation}
    \alpha = \frac{\log NT}{ \omega (1-\gamma) pT}, \hspace{2mm} \textrm{and} \hspace{2mm} T \geq \frac{ 2 C_0 N \tau \zeta' \log NT}{  \omega^2 (1-\gamma)^2 p}.
\label{eqn: choices}
\end{equation}

To simplify the first term in Eq.~\eqref{eqn:main_boundRec}, we use the fact that for all $x\in (0,1)$, it holds that $(1-x) \leq e^{-x}$. Using this in conjunction with the choice of $\alpha$ in \eqref{eqn: choices} yields the following bound on $T_1$ in Eq.~\eqref{eqn:main_boundRec}:

$$ T_1 = O\left( \frac{\max\{\delta^2_0, \sigma^2\}}{NT}\right).$$

To bound $T_2$, we simply substitute the choice of $\alpha$ in Eq.~\eqref{eqn: choices}. For $T_3$, we first substitute the choice of $\alpha$ to obtain:

$$ T_3 = \frac{C_3\tau\sigma^2 {(\log NT)}^3}{\omega^4 (1-\gamma)^4 p^3 T^3}.$$ 
From our choice of $T$ in Eq.~\eqref{eqn: choices}, the following hold:
$$ \frac{\tau \log(NT)}{p \omega^2 (1-\gamma)^2 T} \leq 1, \hspace{3mm}  \frac{N \log(NT)}{Tp} \leq 1.$$
Using these two inequalities, we immediately note that:
$$T_3 = O\left(\frac{\sigma^2 \log(NT)}{p \omega^2 (1-\gamma)^2 NT}\right).$$

Combining the individual bounds on $T_1, T_2$, and $T_3$ leads to Eq.~\eqref{eqn: simplified bound}. Let us complete our derivation with a couple of other points. First, straightforward calculations suffice to check that the choice of $\alpha$ and $T$ in Eq.~\eqref{eqn: choices} meet the requirement on $\alpha$ in the statement of Theorem~\ref{th:main}. Finally, recall from the discussion following Definition~\ref{def:mix} that the mixing time $\tau_{\epsilon}$ satisfies:
$$ \tau_{\epsilon} \leq K \log(1/\epsilon),$$
for some constant $K \geq 1$. Throughout our analysis, we set $\epsilon = \alpha^2$, and then dropped the dependence of $\tau$ on $\epsilon$ for notational convenience. Plugging in the choice of $\alpha$ from Eq.~\eqref{eqn: choices}, we obtain:
$$ \tau \leq 2K \log \left(  \frac{\omega (1-\gamma) pT}{\log(NT)}\right) \leq 2K \log \left(  {\omega (1-\gamma) pT}\right),$$
for $NT \geq e.$ The point of the above calculation is to explicitly demonstrate that one can indeed meet the requirement on $T$ in Eq.~\eqref{eqn: choices} for large enough $T$.

\begin{figure}[t]
\center
	\includegraphics[width=0.5\columnwidth, trim ={3cm 9.5cm 3cm 9.5cm}, clip]{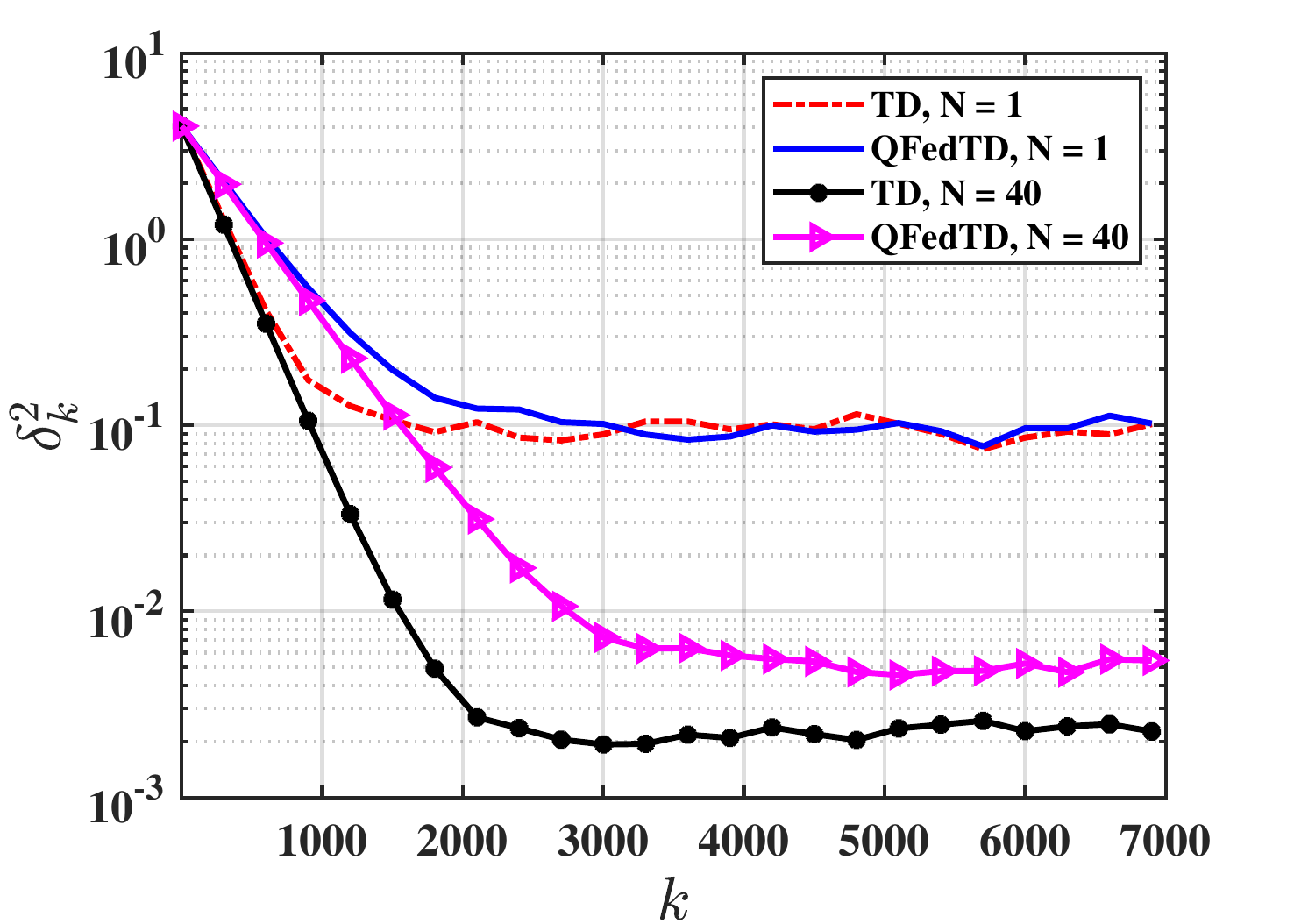}
	\caption{Comparison between vanilla \texttt{FedTD} and \texttt{QFedTD} in single-agent ($N = 1$) and multi-agent ($N = 40$) settings. The number of bits used to quantize the TD update direction is 4 per vector component, and the erasure probability is $p = 0.6$.}
\label{fig:SimConst}
\end{figure}

\begin{figure}[t]
\center
	\includegraphics[width=0.5\columnwidth, trim ={3cm 9.5cm 3cm 9.5cm}, clip]{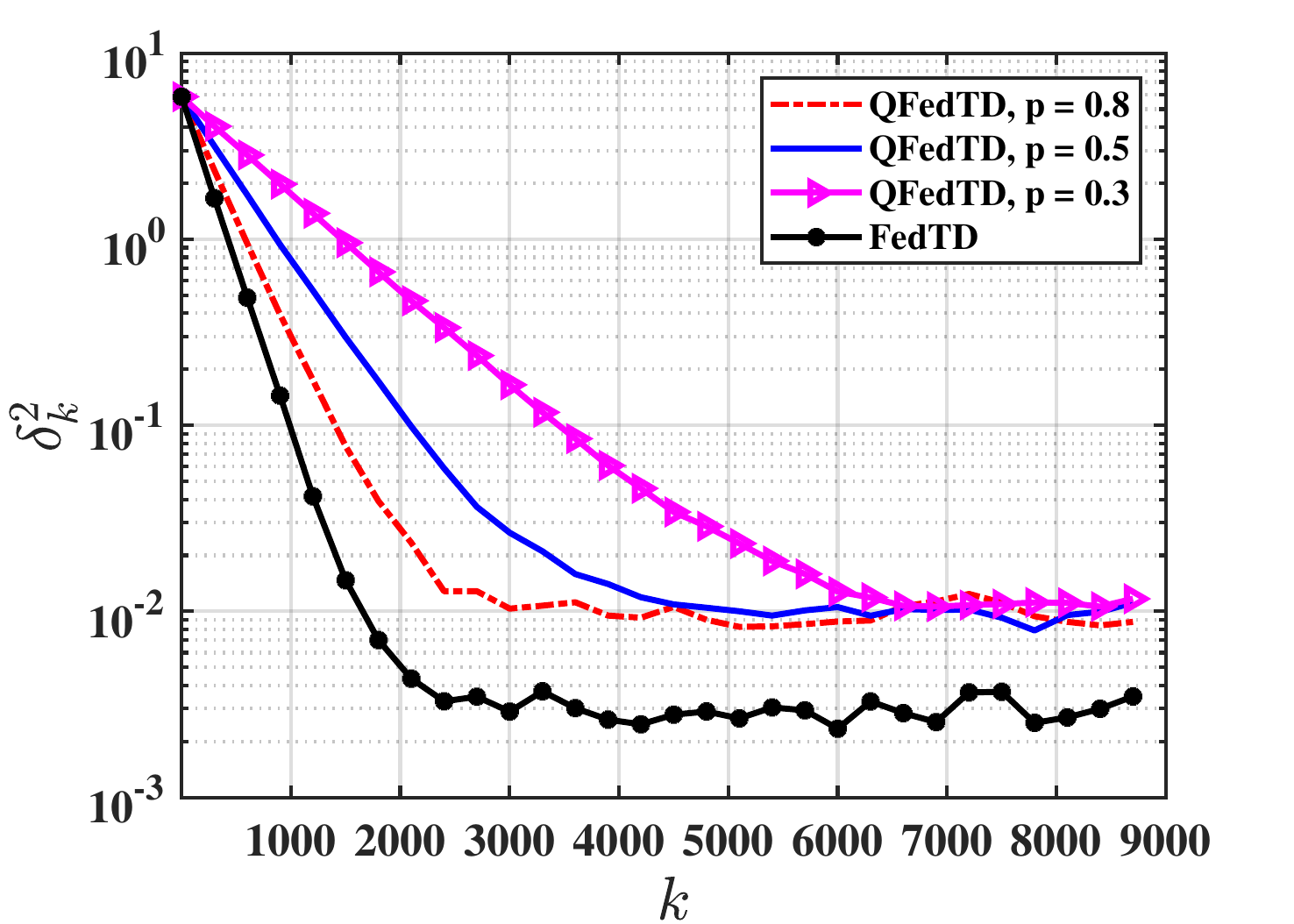}
	\caption{Performance of \texttt{QFedTD} under different values of the erasure probablity $p$. The number of agents cooperating in the simulations performed to obtain this figure is $N = 40$, and each component of the TD update directions is quantized with 4 bits.}
\label{fig:Sim-pComp}
\end{figure}
\begin{figure}[t]
\center
	\includegraphics[width=0.5\columnwidth, trim ={3cm 9.5cm 3cm 9.5cm}, clip]{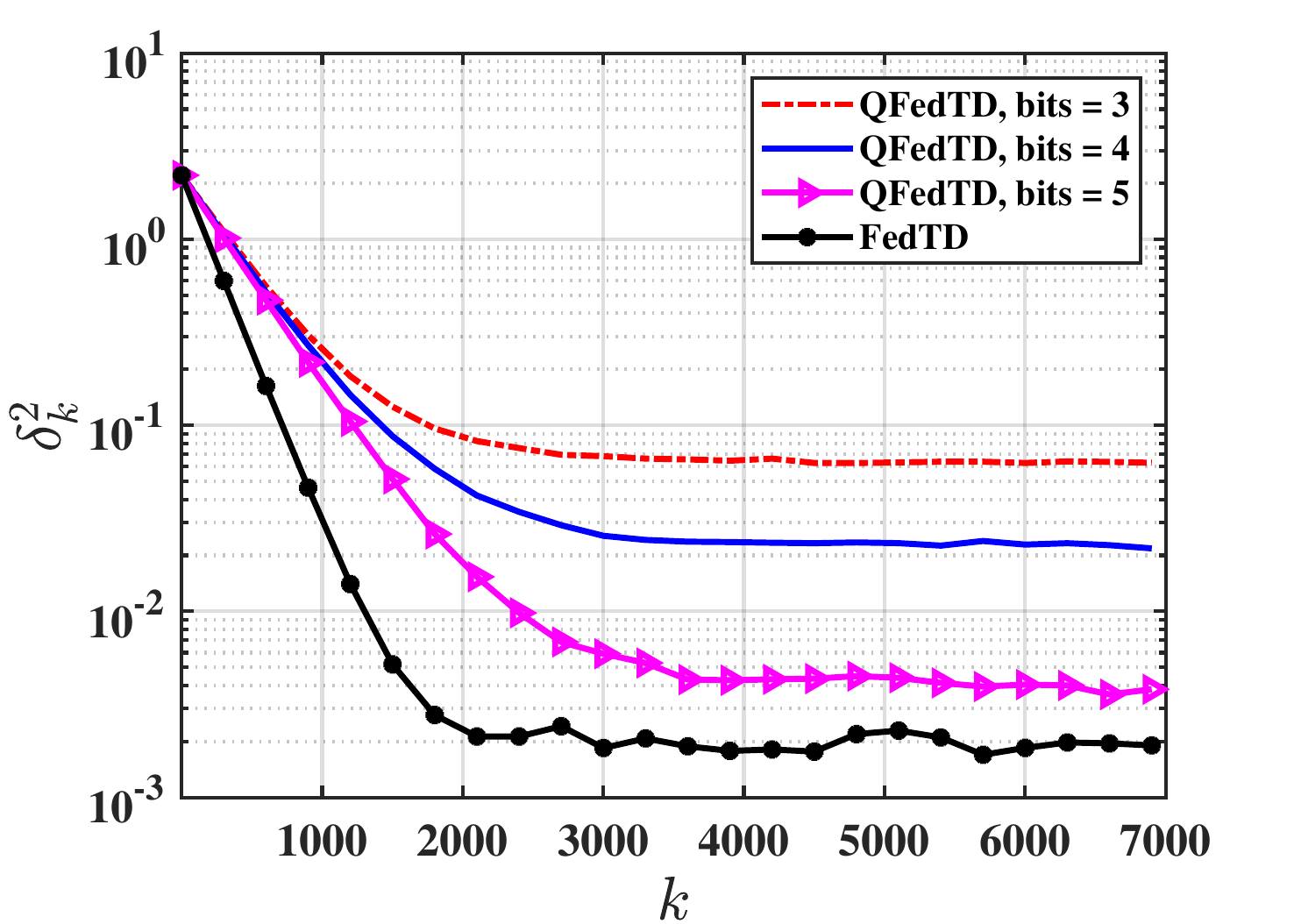}
	\caption{Performance of \texttt{QFedTD} under different values of the number of bits used to quantize the TD update directions. The number of agents cooperating in the simulations performed to obtain this figure is $N = 40$ and the erasure probability is $p = 0.6$.}
\label{fig:Sim-bitsComp}
\end{figure}

\section{Numerical simulations}
\textbf{Basic Setup.} In this section, we provide simulation results on a synthetic example to corroborate our theory. We consider an MDP with $|\mathcal{S}| = 20$ states and a feature space spanned by $d = 10$ orthonormal basis vectors. We fix the discount factor to be $\gamma = 0.5$. In all the simulations, we fix the step size $\alpha = 0.6$ for all the algorithms. In the simulations presented here, we generate the erasure channels with Bernoulli random variables, and employ uniform scalar quantization of the TD update directions, assigning a certain number of bits for the quantization of each vector component of each agent. 

{\textbf{Observation 1: The Linear Speedup Effect.}} In Figure \ref{fig:SimConst}, we compare the proposed \texttt{QFedTD} algorithm with a vanilla version of federated TD learning (i.e., \texttt{QFedTD} without erasures and quantization) that we refer to as \texttt{FedTD}. For the results shown in Figure \ref{fig:SimConst}, we set the probability of successful transmission $p = 0.6$, and we quantize the TD update directions assigning $4$ bits for the quantization of each vector component. In the figure, we show the performance of \texttt{FedTD} and \texttt{QFedTD} in the single agent ($N = 1$) and the multi-agent $(N = 40)$ cases. The simulation results confirm the linear speedup with the number of agents for \texttt{QFedTD}, as established by our theoretical findings. From Figure \ref{fig:SimConst}, we note two important aspects established by the theory and confirmed by the experiment: the rate of convergence of \texttt{QFedTD} is slackened by the probability of unsuccessful transmission $1-p$, while the size of the neighbourhood of $\xs$ to which the algorithm converges is inflated by the quantization noise.

{\textbf{Observation 2: The Effect of the Bernoulli Erasure Channel.}} In Figure \ref{fig:Sim-pComp}, we show the performance of \texttt{QFedTD} under different values of the erasure probability $p$, while fixing the number of agents $N = 40$. Once again, the effect of the successful transmission probability $p$ on the linear convergence rate is evident, consistent with our theoretical result provided in Theorem~\ref{th:main}. Indeed, lowering the probability of successful transmission slows down the rate of convergence; the ball to which the iterates converge has the same size for all the variants.

{\textbf{Observation 3: The Effect of Quantization.}} In Figure \ref{fig:Sim-bitsComp}, similarly to what we did for Figure \ref{fig:Sim-pComp}, we compare the performance of \texttt{QFedTD} for different quantization noise levels. In particular, we show the performance of \texttt{QFedTD} for three values of the number of bits assigned to each vector component of the TD update directions: 3, 4, and 5 bits per vector component. We fix the erasure probability $p = 0.6$, and the number of agents $N = 40$. Consistent with Theorem~\ref{th:main}, we see from this figure that, while the linear rate is the same for all the \texttt{QFedTD} variants, the size of the neighbourhood of $\xs$ to which the algorithm converges is inflated by the quantization noise, i.e., it increases when we diminish the number of bits used to quantize the TD update directions.
\bibliographystyle{IEEEtran}	
\bibliography{IEEEabrv,biblio}

\begin{thebibliography}{10}
\providecommand{\url}[1]{#1}
\csname url@samestyle\endcsname
\providecommand{\newblock}{\relax}
\providecommand{\bibinfo}[2]{#2}
\providecommand{\BIBentrySTDinterwordspacing}{\spaceskip=0pt\relax}
\providecommand{\BIBentryALTinterwordstretchfactor}{4}
\providecommand{\BIBentryALTinterwordspacing}{\spaceskip=\fontdimen2\font plus
\BIBentryALTinterwordstretchfactor\fontdimen3\font minus
  \fontdimen4\font\relax}
\providecommand{\BIBforeignlanguage}[2]{{%
\expandafter\ifx\csname l@#1\endcsname\relax
\typeout{** WARNING: IEEEtran.bst: No hyphenation pattern has been}%
\typeout{** loaded for the language `#1'. Using the pattern for}%
\typeout{** the default language instead.}%
\else
\language=\csname l@#1\endcsname
\fi
#2}}
\providecommand{\BIBdecl}{\relax}
\BIBdecl

\bibitem{konevcny}
J.~Kone{\v{c}}n{\`y}, H.~B. McMahan, D.~Ramage, and P.~Richt{\'a}rik,
  ``{Federated optimization: Distributed machine learning for on-device
  intelligence},'' \emph{arXiv preprint arXiv:1610.02527}, 2016.

\bibitem{mcmahan}
B.~McMahan, E.~Moore, D.~Ramage, S.~Hampson, and B.~A. y~Arcas,
  ``{Communication-efficient learning of deep networks from decentralized
  data},'' in \emph{Artificial Intelligence and Statistics}.\hskip 1em plus
  0.5em minus 0.4em\relax PMLR, 2017, pp. 1273--1282.

\bibitem{bonawitz}
K.~Bonawitz, H.~Eichner, W.~Grieskamp, D.~Huba, A.~Ingerman, V.~Ivanov,
  C.~Kiddon, J.~Kone{\v{c}}n{\`y}, S.~Mazzocchi, H.~B. McMahan \emph{et~al.},
  ``{Towards federated learning at scale: System design},'' \emph{arXiv
  preprint arXiv:1902.01046}, 2019.

\bibitem{stich}
S.~U. Stich, ``{Local SGD converges fast and communicates little},''
  \emph{arXiv preprint arXiv:1805.09767}, 2018.

\bibitem{khaled1}
A.~Khaled, K.~Mishchenko, and P.~Richt{\'a}rik, ``{First analysis of local gd
  on heterogeneous data},'' \emph{arXiv preprint arXiv:1909.04715}, 2019.

\bibitem{khaled2}
------, ``{Tighter theory for local SGD on identical and heterogeneous data},''
  in \emph{International Conference on Artificial Intelligence and
  Statistics}.\hskip 1em plus 0.5em minus 0.4em\relax PMLR, 2020, pp.
  4519--4529.

\bibitem{haddadpour}
F.~Haddadpour and M.~Mahdavi, ``{On the convergence of local descent methods in
  federated learning},'' \emph{arXiv preprint arXiv:1910.14425}, 2019.

\bibitem{woodworth1}
B.~Woodworth, K.~K. Patel, S.~U. Stich, Z.~Dai, B.~Bullins, H.~B. McMahan,
  O.~Shamir, and N.~Srebro, ``{Is Local SGD Better than Minibatch SGD?}''
  \emph{arXiv preprint arXiv:2002.07839}, 2020.

\bibitem{scaffold}
S.~P. Karimireddy, S.~Kale, M.~Mohri, S.~Reddi, S.~Stich, and A.~T. Suresh,
  ``{Scaffold: Stochastic controlled averaging for federated learning},'' in
  \emph{International Conference on Machine Learning}.\hskip 1em plus 0.5em
  minus 0.4em\relax PMLR, 2020, pp. 5132--5143.

\bibitem{acar2021}
D.~A.~E. Acar, Y.~Zhao, R.~M. Navarro, M.~Mattina, P.~N. Whatmough, and
  V.~Saligrama, ``Federated learning based on dynamic regularization,''
  \emph{arXiv preprint arXiv:2111.04263}, 2021.

\bibitem{gorbunov}
E.~Gorbunov, F.~Hanzely, and P.~Richt{\'a}rik, ``{Local SGD: Unified theory and
  new efficient methods},'' in \emph{International Conference on Artificial
  Intelligence and Statistics}.\hskip 1em plus 0.5em minus 0.4em\relax PMLR,
  2021, pp. 3556--3564.

\bibitem{mitraNIPS}
A.~Mitra, R.~Jaafar, G.~J. Pappas, and H.~Hassani, ``{Linear convergence in
  federated learning: Tackling client heterogeneity and sparse gradients},''
  \emph{Advances in Neural Information Processing Systems}, vol.~34, pp.
  14\,606--14\,619, 2021.

\bibitem{proxskip}
K.~Mishchenko, G.~Malinovsky, S.~Stich, and P.~Richt{\'a}rik, ``{ProxSkip: Yes!
  Local Gradient Steps Provably Lead to Communication Acceleration! Finally!}''
  \emph{arXiv preprint arXiv:2202.09357}, 2022.

\bibitem{collinsfedavg}
L.~Collins, H.~Hassani, A.~Mokhtari, and S.~Shakkottai, ``Fedavg with fine
  tuning: Local updates lead to representation learning,'' in \emph{{Advances
  in Neural Information Processing Systems}}.

\bibitem{FedPAQ}
A.~Reisizadeh, A.~Mokhtari, H.~Hassani, A.~Jadbabaie, and R.~Pedarsani,
  ``{Fedpaq: A communication-efficient federated learning method with periodic
  averaging and quantization},'' in \emph{AISTATS}.\hskip 1em plus 0.5em minus
  0.4em\relax PMLR, 2020, pp. 2021--2031.

\bibitem{had21}
F.~Haddadpour, M.~M. Kamani, A.~Mokhtari, and M.~Mahdavi, ``Federated learning
  with compression: Unified analysis and sharp guarantees,'' in
  \emph{International Conference on Artificial Intelligence and
  Statistics}.\hskip 1em plus 0.5em minus 0.4em\relax PMLR, 2021, pp.
  2350--2358.

\bibitem{sutton1988learning}
R.~S. Sutton, ``{Learning to predict by the methods of temporal differences},''
  \emph{Machine learning}, vol.~3, no.~1, pp. 9--44, 1988.

\bibitem{tsitsiklisroy}
J.~N. Tsitsiklis and B.~Van~Roy, ``{An analysis of temporal-difference learning
  with function approximation},'' in \emph{IEEE Transactions on Automatic
  Control}, 1997.

\bibitem{qiFRL}
J.~Qi, Q.~Zhou, L.~Lei, and K.~Zheng, ``{Federated reinforcement learning:
  techniques, applications, and open challenges},'' \emph{arXiv preprint
  arXiv:2108.11887}, 2021.

\bibitem{bhandari2018}
J.~Bhandari, D.~Russo, and R.~Singal, ``A finite time analysis of temporal
  difference learning with linear function approximation,'' in \emph{Conference
  on learning theory}.\hskip 1em plus 0.5em minus 0.4em\relax PMLR, 2018, pp.
  1691--1692.

\bibitem{srikant2019finite}
R.~Srikant and L.~Ying, ``Finite-time error bounds for linear stochastic
  approximation and {TD} learning,'' in \emph{Conference on Learning
  Theory}.\hskip 1em plus 0.5em minus 0.4em\relax PMLR, 2019, pp. 2803--2830.

\bibitem{chenQ}
Z.~Chen, S.~Zhang, T.~T. Doan, S.~T. Maguluri, and J.-P. Clarke, ``Performance
  of q-learning with linear function approximation: Stability and finite-time
  analysis,'' \emph{arXiv preprint arXiv:1905.11425}, p.~4, 2019.

\bibitem{patil2023}
G.~Patil, L.~Prashanth, D.~Nagaraj, and D.~Precup, ``Finite time analysis of
  temporal difference learning with linear function approximation: Tail
  averaging and regularisation,'' in \emph{International Conference on
  Artificial Intelligence and Statistics}.\hskip 1em plus 0.5em minus
  0.4em\relax PMLR, 2023, pp. 5438--5448.

\bibitem{lakshmi}
C.~Lakshminarayanan and C.~Szepesv{\'a}ri, ``{Linear stochastic approximation:
  Constant step-size and iterate averaging},'' \emph{arXiv preprint
  arXiv:1709.04073}, 2017.

\bibitem{dalal}
G.~Dalal, B.~Sz{\"o}r{\'e}nyi, G.~Thoppe, and S.~Mannor, ``{Finite sample
  analyses for TD (0) with function approximation},'' in \emph{Proceedings of
  the AAAI Conference on Artificial Intelligence}, vol.~32, no.~1, 2018.

\bibitem{borkarode}
V.~S. Borkar and S.~P. Meyn, ``The ode method for convergence of stochastic
  approximation and reinforcement learning,'' \emph{SIAM Journal on Control and
  Optimization}, vol.~38, no.~2, pp. 447--469, 2000.

\bibitem{doan}
T.~Doan, S.~Maguluri, and J.~Romberg, ``{Finite-time analysis of distributed TD
  (0) with linear function approximation on multi-agent reinforcement
  learning},'' in \emph{International Conference on Machine Learning}.\hskip
  1em plus 0.5em minus 0.4em\relax PMLR, 2019, pp. 1626--1635.

\bibitem{liuMARL}
R.~Liu and A.~Olshevsky, ``{Distributed TD (0) with almost no communication},''
  \emph{arXiv preprint arXiv:2104.07855}, 2021.

\bibitem{khodadadian2022federated}
S.~Khodadadian, P.~Sharma, G.~Joshi, and S.~T. Maguluri, ``Federated
  reinforcement learning: Linear speedup under markovian sampling,'' in
  \emph{ICML}.\hskip 1em plus 0.5em minus 0.4em\relax PMLR, 2022, pp.
  10\,997--11\,057.

\bibitem{han}
H.~Wang, A.~Mitra, H.~Hassani, G.~J. Pappas, and J.~Anderson, ``Federated
  temporal difference learning with linear function approximation under
  environmental heterogeneity,'' \emph{arXiv:2302.02212}, 2023.

\bibitem{hadjicostis}
C.~N. Hadjicostis and R.~Touri, ``Feedback control utilizing packet dropping
  network links,'' in \emph{Proceedings of the 41st IEEE Conference on Decision
  and Control, 2002.}, vol.~2.\hskip 1em plus 0.5em minus 0.4em\relax IEEE,
  2002, pp. 1205--1210.

\bibitem{schenato}
L.~Schenato, B.~Sinopoli, M.~Franceschetti, K.~Poolla, and S.~S. Sastry,
  ``Foundations of control and estimation over lossy networks,''
  \emph{Proceedings of the IEEE}, vol.~95, no.~1, pp. 163--187, 2007.

\bibitem{rabbat}
M.~G. Rabbat and R.~D. Nowak, ``Quantized incremental algorithms for
  distributed optimization,'' \emph{IEEE Journal on Selected Areas in
  Communications}, vol.~23, no.~4, pp. 798--808, 2005.

\bibitem{doanquant}
T.~T. Doan, S.~T. Maguluri, and J.~Romberg, ``Fast convergence rates of
  distributed subgradient methods with adaptive quantization,'' \emph{IEEE
  Transactions on Automatic Control}, vol.~66, no.~5, pp. 2191--2205, 2020.

\bibitem{reisizadehTSP}
A.~Reisizadeh, A.~Mokhtari, H.~Hassani, and R.~Pedarsani, ``An exact quantized
  decentralized gradient descent algorithm,'' \emph{IEEE Transactions on Signal
  Processing}, vol.~67, no.~19, pp. 4934--4947, 2019.

\bibitem{michelusi}
N.~Michelusi, G.~Scutari, and C.-S. Lee, ``Finite-bit quantization for
  distributed algorithms with linear convergence,'' \emph{IEEE Transactions on
  Information Theory}, vol.~68, no.~11, pp. 7254--7280, 2022.

\bibitem{charles}
Z.~Charles and J.~Kone{\v{c}}n{\`y}, ``{On the outsized importance of learning
  rates in local update methods},'' \emph{arXiv preprint arXiv:2007.00878},
  2020.

\bibitem{nagaraj}
D.~Nagaraj, X.~Wu, G.~Bresler, P.~Jain, and P.~Netrapalli, ``Least squares
  regression with markovian data: Fundamental limits and algorithms,''
  \emph{Advances in neural information processing systems}, vol.~33, pp.
  16\,666--16\,676, 2020.

\bibitem{beznosikov}
A.~Beznosikov, S.~Horv{\'a}th, P.~Richt{\'a}rik, and M.~Safaryan, ``On biased
  compression for distributed learning,'' \emph{arXiv:2002.12410}, 2020.

\bibitem{mitra2023temporal}
A.~Mitra, G.~J. Pappas, and H.~Hassani, ``Temporal difference learning with
  compressed updates: Error-feedback meets reinforcement learning,''
  \emph{arXiv preprint arXiv:2301.00944}, 2023.

\bibitem{levin2017markov}
D.~A. Levin and Y.~Peres, \emph{{Markov chains and mixing times}}.\hskip 1em
  plus 0.5em minus 0.4em\relax American Mathematical Soc., 2017, vol. 107.

\end{thebibliography}

\end{document}